%% file: dataopsy.tex
\documentclass[preprint,journal]{vgtc}       

\graphicspath{{figs/}{figures/}{pictures/}{images/}{./}}

\usepackage{mathptmx}
\usepackage{amsfonts}
\usepackage{fontawesome}
\usepackage{amsmath}

\usepackage{flushend}

\preprinttext{This is the author’s version of the article that has been accepted for publication in IEEE Transactions on Visualization and Computer Graphics.}

\onlineid{0}

\vgtccategory{Research}

\vgtcpapertype{algorithm/technique}

\title{Dataopsy: Scalable and Fluid Visual Exploration using\\ Aggregate Query Sculpting}

\author{
    \authororcid{Md Naimul Hoque}{0000-0003-0878-501X} and
    \authororcid{Niklas Elmqvist}{0000-0001-5805-5301}
}

\authorfooter{
\item Md Naimul Hoque is with University of Maryland, College Park in College Park, MD, USA. E-mail: nhoque@umd.edu.
\item Niklas Elmqvist is with Aarhus University in Aarhus, Denmark; the work was conducted at University of Maryland, College Park, MD, United States. E-mail: elm@cs.au.dk. 
}

\shortauthortitle{Dataopsy}

\abstract{%
    We present \textit{aggregate query sculpting} (AQS), a faceted visual query technique for large-scale multidimensional data.
    As a ``born scalable'' query technique, AQS starts visualization with a single visual mark representing an aggregation of the entire dataset.
    The user can then progressively explore the dataset through a sequence of operations abbreviated as $\mathbb{P}^6$: \textit{pivot} (facet an aggregate based on an attribute), \textit{partition} (lay out a facet in space), \textit{peek} (see inside a subset using an aggregate visual representation), \textit{pile} (merge two or more subsets), \textit{project} (extracting a subset into a new substrate), and \textit{prune} (discard an aggregate not currently of interest).
    We validate AQS with \textsc{Dataopsy}, a prototype implementation of AQS that has been designed for fluid interaction on desktop and touch-based mobile devices.
    We demonstrate AQS and Dataopsy using two case studies and three application examples.
}


\teaser{
  \includegraphics[width=\textwidth]{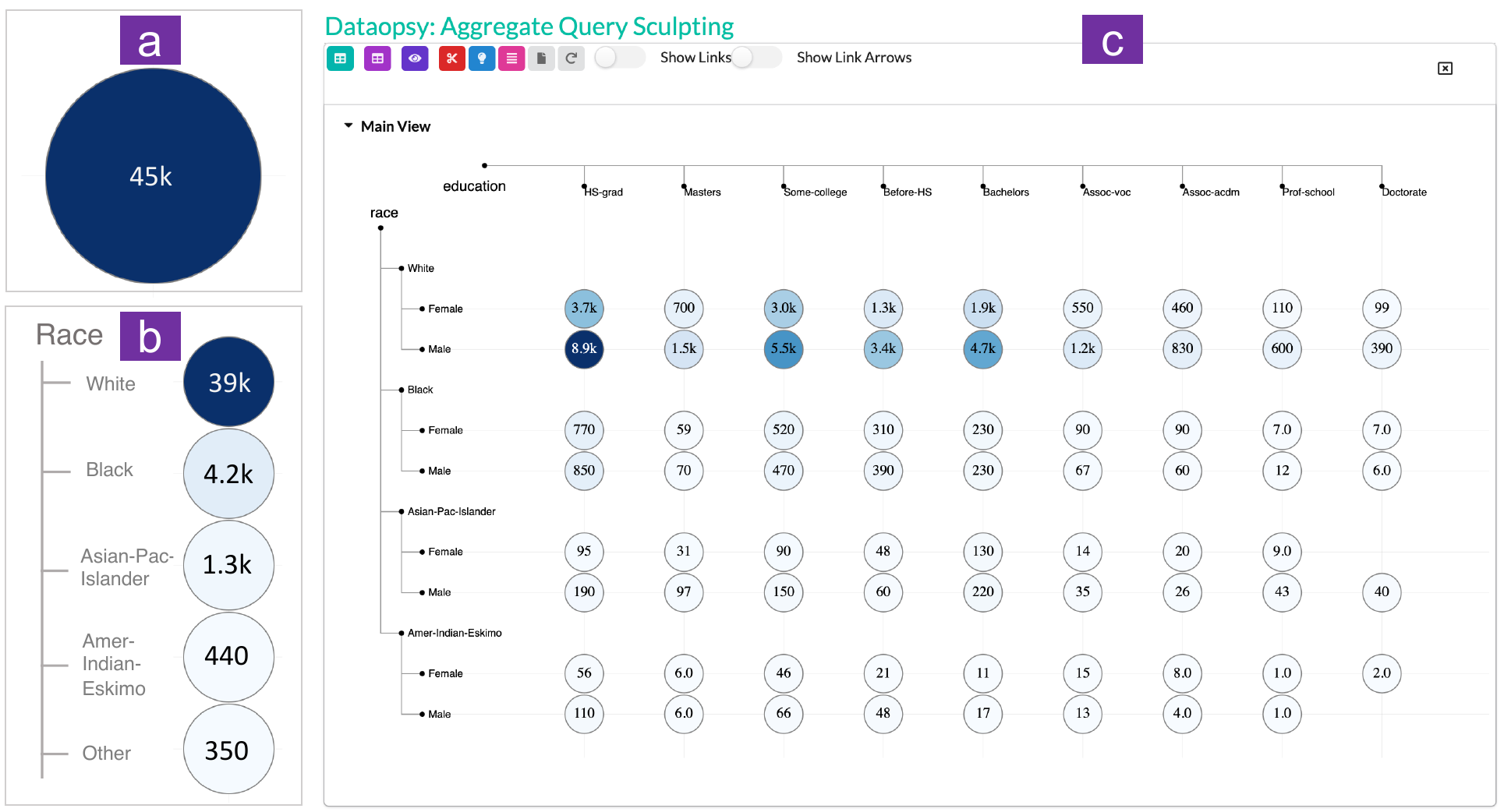}
  \caption{\textbf{The Adult Income~\cite{DBLP:conf/cikm/GhaiMM22} dataset visualized in Dataopsy.}
  (a) The initial view in Dataopsy is a single supernode, an aggregate representation of all 45,000 data points in the dataset.
  Dataopsy provides six interactions ($\mathbb{P}^6$) to explore the dataset iteratively.
  (b) For example, a user can \textit{partition} the initial supernode by \textsc{Race}, transforming it into five new supernodes.
  (c) The user can \textit{prune} the \textsc{Other} category in Race and then add \textsc{Gender} on the vertical axis and \textsc{Education} on the horizontal axis for \textit{partitioning}.
  At the top of each view in Dataopsy, a control panel contains stylized buttons for performing and tracking the interactions.}
  \label{fig:teaser}
}

\begin{document}

\maketitle

\input{content/01-intro}

\input{content/02-background}

\input{content/03-design}

\input{content/04-technique}

\input{content/05-examples}

\input{content/06-discussion}

\input{content/07-conclusion}

\section{Acknowledgments}
This work was partially supported by the U.S. National Science Foundation grant 2211628. Any opinions, findings, and conclusions, or recommendations expressed here are those of the authors and do not necessarily reflect the views of the funding agencies.

\bibliographystyle{abbrv-doi-hyperref}
\bibliography{dataopsy}

\end{document}

%% file: content/01-intro.tex
\section{Introduction}

\textit{E pluribus unum}---Latin for ``out of many, one''---is one of the official mottos of the United States, and also happens to be a dominant strategy for managing scale in data visualization: through \textit{aggregation} of many data items into one visual mark~\cite{DBLP:journals/tvcg/ElmqvistF10}.
Visualizing today's real-world datasets, from the Facebook social network to the billion-parameter large language models (LLMs) of Jurassic-1 and GPT-3, is more or less impractical when using a representation that insists on using one mark per data item---so-called \textit{unit visualizations}~\cite{Park2018b}.
This fact is exacerbated by today's trend towards mobility in data analysis~\cite{Lee2021} using mobile devices that have pathologically small mobile screens~\cite{DBLP:journals/computer/Harrison10}.
And finally, even if we had enough pixels, there is a limit to the number of data items that the human perceptual system can interpret effectively~\cite{DBLP:journals/tvcg/ElmqvistF10, Munzner2014}.
What is truly needed to tackle the next generation of data visualization challenges are techniques that are ``born scalable''; i.e., that have been designed to be scalable from inception.

In this paper, we propose such a ``born scalable'' visualization technique that we call \textit{aggregate query sculpting} (AQS).
AQS is primarily designed for multidimensional tabular data, but can also be used for multivariate networks (i.e., links connecting data items or nodes).
Unlike typical overviews containing thousands or tens of thousands of data items, the initial AQS view is a substrate containing a single \textit{supernode} (aggregate node) representing all of the data items or nodes in the dataset (Figure~\ref{fig:teaser}a).
From this point on, the goal of AQS is to provide the user with a set of fluid interactions to split the supernode into smaller data subsets during exploration, similar to how a sculptor will iteratively cut a large piece of clay into pieces to shape and mold separately.
There are six such interactions, and they are summarized by the acronym $\mathbb{P}^6$ for \textit{pivot}, \textit{partition}, \textit{peek}, \textit{pile}, \textit{project}, and \textit{prune}.
Pivoting splits a supernode into facet supernodes based on a data attribute; partitioning lays out facet nodes in visual space; peeking shows the data distribution inside a node; piling merges two or more nodes; projection extracts a subset into a new visual substrate where $\mathbb{P}^6$ operations can continue; and pruning eliminates undesired supernodes.

To validate the utility of aggregate query sculpting, we present \textsc{Dataopsy}, a web-based prototype implementation of AQS capable of visualizing thousands of entities in a standard web browser.
Dataopsy and its AQS concepts draw on many existing concepts.
Pivoting is based on graph pivoting techniques from PivotGraph~\cite{DBLP:conf/chi/Wattenberg06} as well as Tableau (n{\'e}e Polaris~\cite{DBLP:conf/infovis/StolteH00}).
Microsoft's Sand Dance~\cite{Drucker2015} and Google's Facets~\cite{Facets2017, Wexler2017} use a similar attribute-based 2D layout, but Dataopsy is designed for networks and maintains aggregate supernodes instead of a unit representation. 
Furthermore, the projection technique, which is inspired by Shneiderman and Aris's semantic substrates~\cite{DBLP:journals/ivs/ArisS07, DBLP:journals/tvcg/ShneidermanA06}, enables creating multiple linked substrates in the same visual space to avoid inelegant deep facet nesting, which is a problem for PivotGraph, Polaris, Sand Dance, and Facet browsers.
Finally, the Dataopsy interface is designed for fluid interaction on a desktop and tablet.
The result is a smooth and scalable data exploration application for multivariate data synthesizing features no comparable tool possesses.

We present two case studies and three application examples involving Dataopsy to demonstrate the utility of aggregate query sculpting.
In the first case study, two data scientists and algorithmic fairness researchers used Dataopsy to evaluate machine bias in the Adult Income dataset~\cite{DBLP:conf/cikm/GhaiMM22, DBLP:conf/ieeevast/CabreraEHKMC19}. In the second case study, a creative writer used Dataopsy to navigate a complex set of scenes, locations, chapters, characters, and events from a fiction novel and sculpted it for adaptation to a screenplay format. As an application example, we showed how an analyst could use Dataopsy to understand the linguistic properties of inter-community conflicts from 300,000 Reddit posts~\cite{DBLP:conf/www/KumarHLJ18}.
We then analyzed 1.7 billion taxi rides in New York City to identify hotspots for rides~\cite{NYCTaxi}.
Finally, we used the VisPub~\cite{DBLP:journals/tvcg/IsenbergHKIXSSC17} dataset to analyze the IEEE VIS scientific community over the years.
Overall, the case studies and examples show the generalizability and scalability of AQS and Dataopsy in exploring diverse multidimensional datasets.

%% file: content/02-background.tex
\section{Background}

Building and managing queries is as old as visualization itself; the ``zoom and filter'' part of Shneiderman's visual information seeking mantra~\cite{DBLP:conf/vl/Shneiderman96} refers to controlling which data items to show on the screen, primarily to manage scale. 
In this section, we review the related work on query management and visual information seeking, including for multivariate datasets, faceted browsing, and multivariate graphs.

\subsection{Multivariate or Multidimensional Visual Exploration}

Multidimensional datasets consist of many attributes per observation, and are routinely found in both tabular as well network applications. 
For example, the U.S.\ Census dataset of citizen demographics includes hundreds of attributes capturing individuals living in the United States, including properties such as age, gender, education, annual income, and marital status. 
Searching and filtering such multivariate datasets was a challenging prospect often involving writing SQL queries until Williamson and Shneiderman proposed \textit{dynamic queries} using double-ended range sliders~\cite{DBLP:conf/sigir/WilliamsonS92}.
These sliders enabled selecting an interval in both quantitative and---later---categorical axes~\cite{DBLP:conf/chi/AhlbergS94d} in the dataset.

Significant work has since been conducted on searching and querying multidimensional datasets.
Many visual query techniques are intimately tied to a visual representation.
For example, axis filtering~\cite{DBLP:journals/computer/SeoS02} is designed for filtering on the axes in a parallel coordinate plot.
ExPlates~\cite{DBLP:journals/cgf/JavedE13} \textit{spatializes} multidimensional interaction into 2D space.
As the name implies, the ScatterDice system~\cite{DBLP:journals/tvcg/ElmqvistDF08} is based on scatterplot matrices and introduces the concept of \textit{query sculpting} where a dataset is filtered from different angles until the final desired result is reached.
The idea was later generalized in the VisDock~\cite{DBLP:journals/tvcg/ChoiPWFE15} cross-cutting interaction library and became a fundamental feature of the Keshif visual data browser~\cite{DBLP:journals/tvcg/YalcinEB16, DBLP:journals/tvcg/YalcinEB18}.
In this paper, we build on query sculpting but generalize it to visual aggregates, where massive datasets have been hierarchically grouped into aggregation trees for scalability~\cite{DBLP:journals/tvcg/ElmqvistF10}.

Polaris~\cite{DBLP:conf/infovis/StolteH00} and FromDaDy~\cite{DBLP:journals/tvcg/HurterTC09} were early examples of highly interactive multivariate query and visualization systems.
Most multivariate visualizations are \textit{unit visualizations}~\cite{Park2018b} in that they represent each data item with exactly one visual mark.
More recently, Microsoft's SandDance~\cite{Drucker2015} and Google Facets~\cite{Facets2017, Wexler2017} enable using similar interaction, layout, and query techniques to visualize multivariate datasets, such as machine learning training data.
ATOM~\cite{Park2018b} was designed as a declarative grammar for building unit visualizations. We differ from prior works in several dimensions.
\textit{First,} instead of unit marks, we use aggregated marks to scale unit visualization to large datasets.
\textit{Second,} AQS introduces six interactions for iterative explorations of the aggregated marks. 
While some of these interactions are motivated by prior works, the combination of them provides analytical capabilities that no prior works possess.
For example, Polaris's Cross and Nest operations motivated our pivot and partitioning operations. 
However, Cross and Nest could create inelegant nesting and visual clutter.
We solve this limitation by integrating the Projection operation, motivated from semantic subtrates~\cite{DBLP:journals/ivs/ArisS07, DBLP:journals/tvcg/ShneidermanA06}. 
\textit{Finally,} prior works primarily focus on desktop applications whereas AQS is suitable for data analysis in touch-based mobile devices and extends to network analysis.

\subsection{Faceted Browsing}

Faceted browsing~\cite{DBLP:conf/chi/YeeSLH03}, where a corpus is explored along one or more conceptual dimensions, was introduced as an alternative to keyword search and image similarity for browsing large-scale image repositories.
The idea was quickly generalized to any multidimensional dataset and then adopted by many internet search providers, particularly for e-commerce and real estate websites. 

Of course, faceted browsing is a powerful idea with applications to many information retrieval and query research problems.
One of the early applications was FacetMap~\cite{DBLP:journals/tvcg/SmithCMRRT06}, a highly visual and dynamic visualization that summarizes the current state of the filters and search results based on a space-filling rectilinear layout approach.
FacetMap shares many similarities with Dataopsy and AQS, but our approach uses a single set of vertical and horizontal axis mappings for displaying dimensions. 
For this reason, Dataopsy yields visual representations that are more stable and easier to understand.
Nevertheless, we draw inspiration from FacetMap's aggregated and scalable visual encoding.

Several other research tools are based on faceted browsing.
FacetLens~\cite{DBLP:conf/chi/LeeSRCT09} build on FacetMap and uses a similar representation, but support visual comparison view as well as more advanced pivoting operations.
FacetZoom~\cite{DBLP:conf/chi/DachseltFW08} enable smooth exploration of hierarchical metadata using a continuous zooming interaction.
Finally, PivotPaths~\cite{DBLP:journals/tvcg/DorkRRD12} provides a fluid and highly interactive browsing experience that externalizes the links (or paths) between different facet values.
These existing tools all served as inspiration for our work in this paper.

\subsection{Multivariate Graphs}

From a data visualization perspective, the leap from table to network is small: all you need are relations connecting entities~\cite{Munzner2014}.
Practically, this means adding a second ``edge table'' linking keys in the original node table.
While layout techniques are mostly radically different for tables vs.\ networks, there is one approach that is shared: \textit{attribute-based layout}~\cite{DBLP:conf/chi/Wattenberg06, DBLP:journals/cgf/NobreMSL19}, where the position of a node on a geometric axis is dependent on a specific attribute associated with the node.
Not surprisingly, this kind of visual mapping is commonly applied to data points when mapping a data table to visual space, such as in a scatterplot.

A canonical example of attribute-based layout is PivotGraphs~\cite{DBLP:conf/chi/Wattenberg06}, which uses vertical and horizontal space to unpack a single aggregated node in a node-link into 2D space.
Aggregated edges show relations in the resulting graph.
Our work in this paper draws heavily on PivotGraphs, but generalizes the idea to both tabular as well as network data, and also introduces several new operations to improve on the idea.

GraphDice~\cite{DBLP:journals/cgf/BezerianosCDEF10} is another example of attribute-based layout: it is essentially a network version of the original ScatterDice~\cite{DBLP:journals/tvcg/ElmqvistDF08} system discussed above.
However, unlike PivotGraphs, GraphDice is a unit visualization system with one visual mark per data item.
While our Dataopsy system is based on visual aggregates, we borrow the faceted navigation supported by the GraphDice tool for our work.

Scale is a perpetual problem for graph visualization.
ASK-GraphView~\cite{DBLP:journals/tvcg/AbelloHK06} supports interactive visual exploration of node-link diagrams consisting of millions of nodes through the use of clustering and animation.
ZAME~\cite{DBLP:conf/apvis/ElmqvistDGHF08} instead uses adjacency matrices and a level-of-detail pyramid to support massive scale.
DOI graphs~\cite{DBLP:journals/tvcg/HamP09} handle the problem by showing only subsets of graphs. 
Finally, Refinery~\cite{DBLP:journals/cgf/KairamRDFH15} uses a similar form of ``associative browsing'' to support browsing on limited neighborhoods of a massive heterogeneous graph.

Heterogeneous---or multimodal---graphs are a specialized subset of multivariate graphs because one attribute governs the \textit{type} of the node; e.g., students and courses in a registrar's database, books, magazines, and digital media in a library database, or gokarts, trainers, and drivers in a racing club roster.
Aris and Shneiderman studied how to best visualize such multimodal data by separating them into individual \textit{semantic substrates}~\cite{DBLP:journals/ivs/ArisS07, DBLP:journals/tvcg/ShneidermanA06}, one for each node type.
Ghani et al.~\cite{DBLP:journals/tvcg/GhaniKLYE13} studied the use of visualization techniques for heterogeneous data for social network analysis, presenting an approach akin to parallel coordinate displays for network data in response.
Finally, Ploceus~\cite{DBLP:journals/ivs/LiuNS14} generalizes the idea of linked tables yielding heterogeneous networks, presenting an algebra and an interactive visualization system to support it.
All of these existing tools and techniques were influential in our design of AQS.
However, none of them provide the same kind of fluid, highly interactive, and scalable approach to visual exploration and querying that AQS and Dataopsy do.

%% file: content/03-design.tex
\section{Aggregate Query Sculpting}

\textit{Aggregate query sculpting} (AQS) is an iterative filtering technique for multivariate data.
It draws inspiration from the ScatterDice technique~\cite{DBLP:journals/tvcg/ElmqvistDF08} where the \textit{query sculpting} concept was introduced based on the metaphor of a sculptor repeatedly chiseling away at a block of stone until the sculpture is complete.
However, while the original implementation of the techniques was intended for unit visualizations~\cite{Park2018b} such as scatterplots, where each individual data point is represented by a unique visual mark, aggregate query sculpting, as the name suggests, is designed for aggregated visual representations where individual marks can represent many---potentially thousands or even millions---of data items.
We use the term ``born scalable'' to refer to visual representations and interaction techniques that were designed for massive scale.

Here we describe the data model for aggregate query sculpting and the fundamental $\mathbb{P}^6$ operations in abstract terms.
In the following section, we discuss how to implement the $\mathbb{P}^6$ operations in the web-based \textsc{Dataopsy} prototype tool for multivariate data.

\subsection{Design Rationale}

We designed aggregate query sculpting by harnessing prior art from the literature with three specific design goals (DG1--DG3) in mind:

\begin{itemize}
    
    \item[DG1]\textbf{Born scalable:} Realistic datasets cannot be visualized as unit visualizations~\cite{Park2018b} because of both technical and perceptual limitations. 
    Instead, robust visual representations must be designed from the ground up using visual and data aggregation~\cite{DBLP:journals/tvcg/ElmqvistF10}.
    
    \item[DG2]\textbf{Fluid interaction:} We envision an iterative and progressive filtering method based on fluid interaction~\cite{DBLP:journals/ivs/ElmqvistMJCRJ11}, where user actions promote flow~\cite{Csikszentmihalyi1991}, are based on direct manipulation~\cite{Shneiderman1983}, and minimize the gulfs of execution and evaluation~\cite{Norman1986}
    
    \item[DG3]\textbf{Faceted browsing:} Multidimensional datasets are easily navigated, filtered, and queried using \textit{faceted search}~\cite{DBLP:conf/chi/YeeSLH03} where filters can be expressed across multiple hierarchical dimensions (\textit{facets}).
    
\end{itemize}

\subsection{Data Model}

All AQS operations are applied to a multidimensional dataset $\mathbb{D}' \subseteq \mathbb{D}$, where $\mathbb{D}$ is the full dataset currently being visualized.
The $\mathbb{D}'$ is called a \textit{supernode} even if the data is not relational. 
Supernodes that are part of networks also have \textit{superlinks} $\mathbb{E}'$; edge subsets drawn from the full edge set $\mathbb{E}$.
Some AQS operations operate purely on a data level, whereas others operate on a visual level, and others still operate on both.
Furthermore, the operations tend to apply either to entire rows or entire columns---or the entire dataset---in a substrate based on the visual layout.
We discuss these details for each operation below.

\subsection{Visual Representation}

A multidimensional visual representation managed using aggregate query sculpting includes one or more \textit{semantic substrates}~\cite{DBLP:journals/ivs/ArisS07, DBLP:journals/tvcg/ShneidermanA06}: a 2D visual space of any geometric dimension.
Additional substrates can be laid out depending on the nature of the data and the user's wishes; for example, two substrates can share a horizontal axis, making it useful to stack the substrates vertically (one above the other).

Each substrate contains one or more supernodes $\mathbb{D}'$ (which could potentially be the entire dataset $\mathbb{D}$ if only one substrate is in use).
Inside the substrate, visual aggregates representing datasets are organized in a regular 2D grid with row and column headers.
Each supernode $\mathbb{D}'$ is represented by a single visual mark (DG1); this is typically a simple 2D geometric shape such as a circle.
The size of the underlying data can be conveyed using multiple different methods (sometimes redundantly), such as using a label, a color scale, or the size of shape. 
When the dataset is a multivariate network, the relationships (edges) between entities (vertices) in the underlying network are also aggregated into superlinks that are represented using single link marks (DG1).
Again, the number of aggregated edges can be conveyed using color or thickness.

\subsection{Interaction}

\begin{figure}[htb]
    \centering
    \subfloat[Row-level operations.]{
        \includegraphics[width=0.49\linewidth]{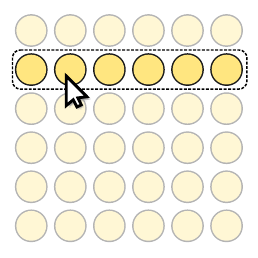}
        \label{fig:aqs-row}
    }
    \subfloat[Column-level operations.]{
        \includegraphics[width=0.49\linewidth]{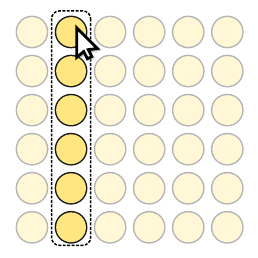}
        \label{fig:aqs-column}
    }
    \caption{\textbf{Operation scope.}
    Aggregate query sculpting operations are performed on entire facets, which means that the scope will be entire rows or columns (or dimensions that are currently hidden).}
    \label{fig:scope}
\end{figure}

We envision aggregate query sculpting as a highly interactive and fluid~\cite{DBLP:journals/ivs/ElmqvistMJCRJ11} query technique where the user rapidly performs multiple operations to identify the data they are interested in (DG2). 
Furthermore, operations can be performed on individual rows or columns, or entire groups of rows or columns by using the partitioning hierarchy (Figure~\ref{fig:scope}).
To facilitate such rapid, direct, and reversible interaction, we suggest including an interaction stack where users can easily overview, undo, and redo individual operations.

\subsection{Query Operations}

We define aggregate query sculpting based on six fundamental sculpting operations that we call $\mathbb{P}^6$ for \textit{pivot}, \textit{partition}, \textit{peek}, \textit{pile}, \textit{project}, and \textit{prune}.
Each operation is applied on a row or column basis to ensure a consistent grid layout.
For the treatment below, assume that $\mathbb{D}$ is a multidimensional dataset consisting of cars with standard dimensions such as gas mileage, acceleration, weight, cylinders, origin, etc.

\paragraph{\faDelicious~Pivot.} 

\begin{figure}[htb]
    \centering
    \subfloat[A single supernode.]{
        \includegraphics[height=2cm]{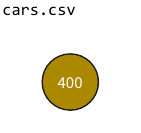}
        \label{fig:aqs-one}
    }   
    \subfloat[Three pivoted supernodes.]{
        \includegraphics[height=2cm]{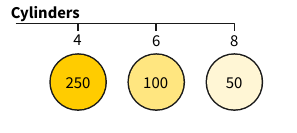}
        \label{fig:aqs-partition}
    }   
    \caption{\textbf{Pivoting and partitioning supernodes.}
    Laying out pivoted supernodes along the horizontal axis.}
    \label{fig:pivot}
\end{figure}

The \textit{pivot} operation splits a supernode $\mathbb{D}'$ into $N$ disjoint supernodes $\{ \mathbb{D}'_1, \ldots, \mathbb{D}'_N \}$ based on a group criterion. 
This is a generalized form of faceted browsing (DG3).
Some group criteria use nominal data types; for example, we could imagine pivoting $\mathbb{D}'$ based on the number of cylinders, resulting in three supernodes $\mathbb{D}'_i$ for 4, 6, and 8 cylinders (Figure~\ref{fig:pivot}).
Alternatively, we could pivot by binning a quantitative value, such as five intervals of gas mileages for $[0, 10), [10, 20), [20, 30), [30, 40)$, and $[40, \infty)$ miles per gallons.

\paragraph{\faTable~Partition.} 

\begin{figure}[htb]
    \centering
    \includegraphics[width=0.75\linewidth]{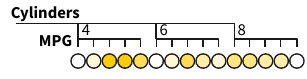}
    \caption{\textbf{Nested axis partitioning.}
    Example showing how to nest multiple pivoted axes inside a single geometric axis.}
    \label{fig:nested-partitioning}
\end{figure}

The \textit{partition} operation is a pure visual operation for laying out supernodes $\mathbb{D}'_i$ in 2D space along a vertical or horizontal geometric axis.
Partitioning will use the entire available space along the chosen geometric axis in the current substrate.
This is typically done by allocating an equal amount of visual space to each supernode, although it is also possible to allocate visual space proportional to the size of each supernode (i.e., the number of items in each supernode).
The approach is similar to PivotGraphs~\cite{DBLP:conf/chi/Wattenberg06} and Polaris~\cite{DBLP:conf/infovis/StolteH00}, but supports nesting in multiple levels. If the chosen geometric axis has already been used for partitioning, the next level of partitioning will be nested.
For example, if we first partition the horizontal axis based on the three sets of cylinders (4, 6, and 8), we can then partition each of these three categories based on the four gas mileage groups; see Figure~\ref{fig:nested-partitioning}.

\paragraph{\faEye~Peek.} 

Sometimes the user wants to see inside a supernode without pivoting and partitioning.
The \textit{peek} operation transforms the visual representation of one or all aggregate marks into a \textit{glyph representation}~\cite{DBLP:journals/tvcg/ElmqvistF10} showing the contents of each mark based on some axis.
For example, peeking can change the color-coded circles into pie charts showing the origin (U.S., Europe, or Asia) of each group of cars pivoted and partitioned based on number of cylinders and then gas mileage.

\begin{figure}[htb]
    \centering
    \includegraphics[width=0.75\linewidth]{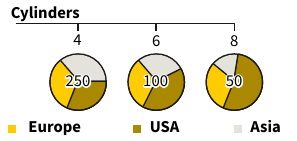}
    \caption{\textbf{Peeking into supernodes.}
    Representing each visual aggregate as a pie chart showing the origin of each subset of cars.}
    \label{fig:peeking}
\end{figure}

\paragraph{\faReorder~Pile.} 

\begin{figure}[htb]
    \centering
    \subfloat[Selecting two supernodes to pile.]{
        \includegraphics[width=0.49\linewidth]{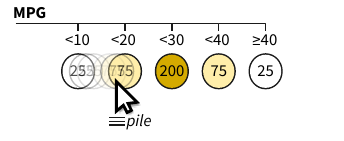}
        \label{fig:aqs-pile1}
    }
    \subfloat[Resulting four supernodes.]{
        \includegraphics[width=0.49\linewidth]{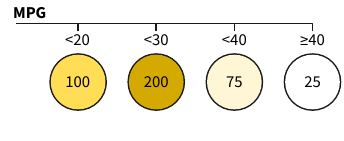}
        \label{fig:aqs-pile2}
    }
    \caption{\textbf{Piling supernodes.}
    Grouping the two lowest mileage supernode of cars into a single supernode.}
    \label{fig:piling}
\end{figure}

\textit{Piling} merges two or more selected supernodes into a single supernode (i.e., as the union of the selected supernodes), potentially enabling the user to name the resulting supernode.
This can be useful when an automatic binning operation yields too many individual supernodes, some of which are meaningless on their own.
For example, the user could choose to pile the [0, 10) and [10, 20) gas mileage supernode into a single supernode that they name ``poor fuel economy'' (Figure~\ref{fig:piling}).

\paragraph{\faLightbulbO~Project.}

\begin{figure}[htb]
    \centering
    \includegraphics[width=\linewidth]{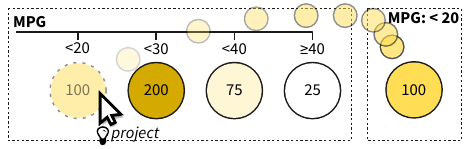}
    \caption{\textbf{Projecting to a new substrate.}
    Extracting low fuel economy cars to a new semantic substrate for continued exploration.}
    \label{fig:projection}
\end{figure}

Partitioning multiple pivots into the same geometric axis will eventually yield deep nesting and an explosion of supernode combinations. 
To reduce clutter, the \textit{project} operation enables selecting a subset of the data and projecting it onto a new semantic substrate that is laid out independent of the originating substrate.
The selected data is subtracted from the original substrate, ensuring that the substrates remain disjoint.
For example, the user could select all of the low fuel economy cars and project them onto a new substrate to enable further exploration while avoiding to add to the existing nested hierarchy of partitions (Figure~\ref{fig:projection}). 

\paragraph{\faScissors~Prune.} 

\begin{figure}[htb]
    \centering
    \subfloat[Selecting a category to prune.]{
        \includegraphics[width=0.33\linewidth]{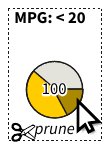}
        \label{fig:aqs-prune1}
    }
    \subfloat[Resulting supernode.]{
        \includegraphics[width=0.33\linewidth]{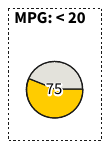}
        \label{fig:aqs-prune2}
    }
    \caption{\textbf{Pruning supernodes.}
    Eliminating entire subsets or values from consideration using pruning.}
    \label{fig:pruning}
\end{figure}

Finally, \textit{prune} allows for eliminating (e.g., hiding; all actions are reversible) selected supernodes from view. 
The operation is similar to the FromDaDy multidimensional visualization tool~\cite{DBLP:journals/tvcg/HurterTC09}.
It can be applied to entire nested hierarchies, or to specific data values. 
For example, the user could easily eliminate all U.S.\ cars from the low fuel economy substrate by pruning on that data value in the origin dimension (Figure~\ref{fig:pruning}).

%% file: content/04-technique.tex
\begin{figure}[hbt]
    \centering
    \includegraphics[width=0.9\columnwidth]{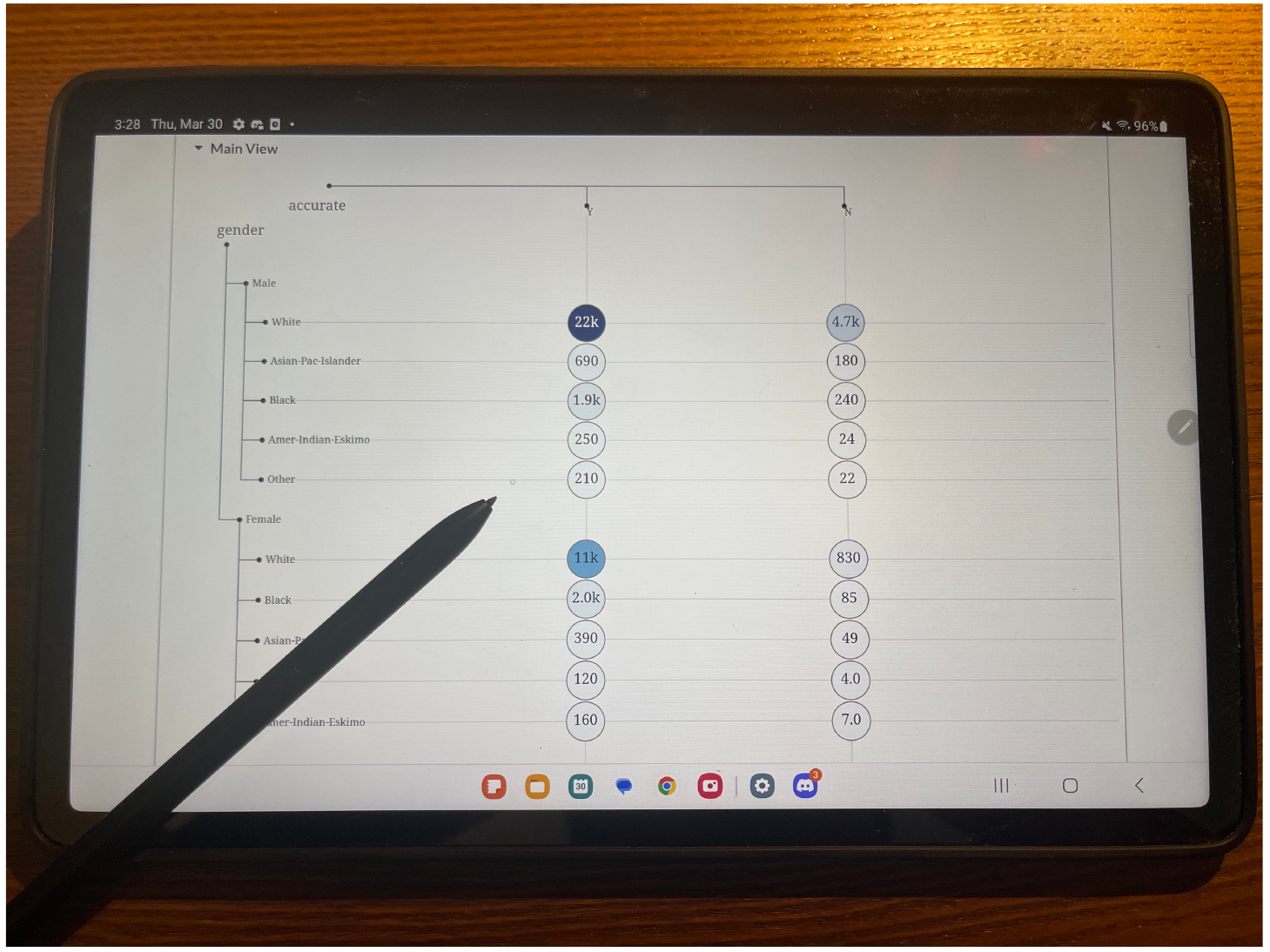}
    \caption{\textbf{Dataopsy on a Samsung Galaxy S8 tablet.}
    The tool is particularly powerful on a mobile tablet device with touch or pen interaction.
    }
    \label{fig:tablet}
\end{figure}


\begin{figure}[h]
    \centering
    \includegraphics[width=0.95\columnwidth]{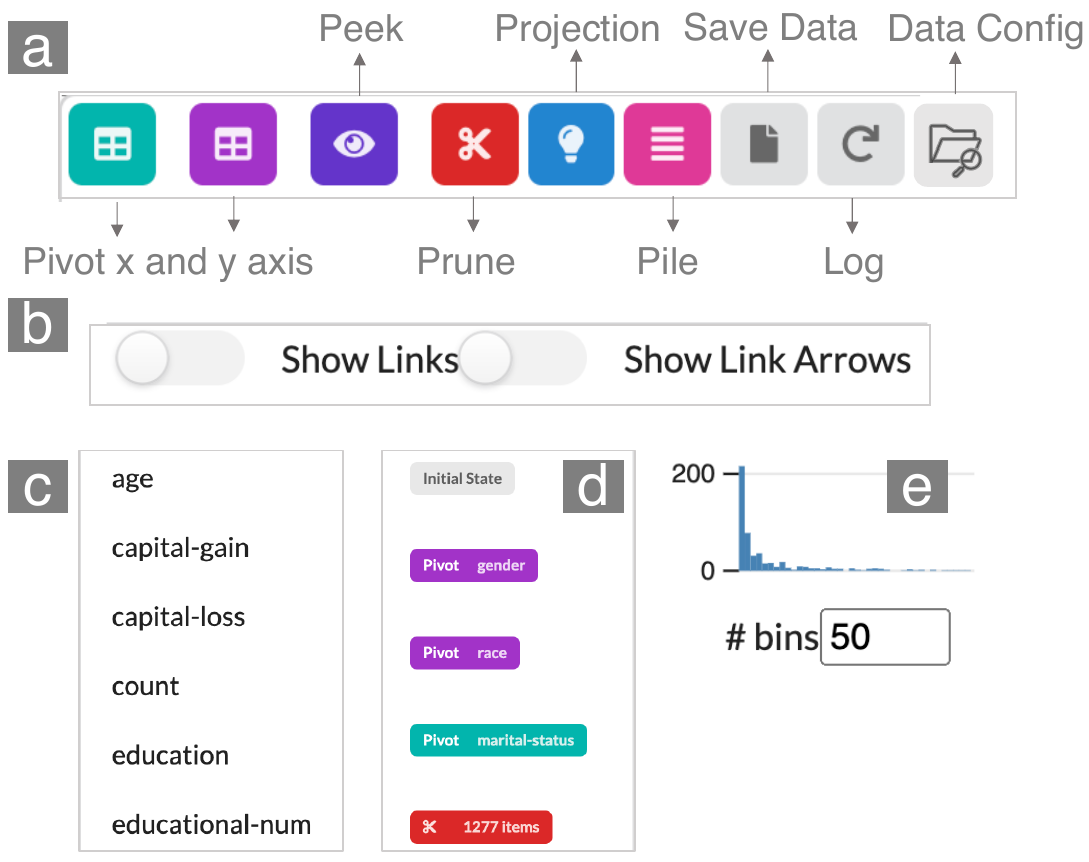}
    \caption{\textbf{Different components of the card header.}
    (a) Buttons and dropdowns as styled icons for supporting $\mathbb{P}^6$.
    (b) Two checkboxes to toggle seeing links and link arrows. (c) A sample dropdown containing the data attributes that opens when a user clicks on either \faTable~partition or \faEye~peek icons.~(d) Interaction log recorded by Dataopsy.
    This dropdown opens up when a user clicks on the Log icon.
    The user can go back and forth between different stages using this dropdown.~(e) A histogram of the target variable.
    A user can select and prune data using this histogram (e.g., pruning data points with less than 5 occurrences).}
    \label{fig:header}
\end{figure}

\section{Dataopsy: AQS for Multidimensional Data}

We developed Dataopsy, a web-based visual analytics tool, to demonstrate AQS in practice.
Dataopsy can be used in a standard web browser using any medium to large screen device
For example, Figure~\ref{fig:tablet} shows Dataopsy on a Samsung Galaxy S8 tablet device.
In this section, we describe the visual interface of Dataopsy as well as query actions and interactions supported.
We also include a video demonstration in the supplemental materials.

\subsection{Card Design}

The central user interface (UI) component of Dataopsy is a card (Figure~\ref{fig:teaser}c), following the design of the popular UI component with the same name.\footnote{\url{https://www.nngroup.com/articles/cards-component/}}
Cards are typically used to couple relevant information into a modular container. 
We chose cards as our core component as our system should support semantic substrates, views with identical functionalities albeit different underlying data subsets.

 The default card, called Main, contains all data points in a single supernode (Figure~\ref{fig:teaser}a).
From this initial card, the user can use the \faLightbulbO~Projection operation to create a new substrate, which is then projected in a new card with identical design.
The cards in our system are flexible.
By default, they align horizontally and take 2/3 of the horizontal and vertical dimensions of the screen; but users can change their order, collapse them, or delete them at any time. Each card has two sub-components: a header and a body. 

\subsubsection{Card Header}
The header contains styled icons to support $\mathbb{P}^6$ (Figure~\ref{fig:header}). We do not include a separate icon for \faDelicious~pivoting as  \faTable~partitioning or laying out the visual marks on the 2D space directly depends on pivoting. Instead, we provide two icons and dropdowns within them to \faTable~partition horizontal and vertical axis. Clicking the \faScissors~prune icon deletes selected data points from the card. Similarly, clicking the \faLightbulbO~project icon copies selected data points from the current card and opens a new card with the copied data. \faReorder~Pile option combines selected data points together. Users can optionally provide a name to the merged categories using a popup. A user selects data points by directly interacting with the visualization (described in Section~\ref{sec:viz_interaction}).

All AQS operations are saved in an \faRepeat~interaction log or stack (Figure~\ref{fig:header}d). Using this stack, a user can go back and forth between any stage of the exploration process. Further, we provide options to \faFile~save and download the current state of the data as a CSV file. This is helpful for exporting data after transforming the original data using pruning and piling. Finally, a user can optionally \faFolder~configure data attributes such as defining alphabetical or numerical sorting options for the attributes.

\subsubsection{Card Body}
The card body contains the SVG container for the visualization. We describe the visualization design within the card body next.


\begin{figure}[tbh]
    \centering
    \includegraphics[width=0.99\columnwidth]{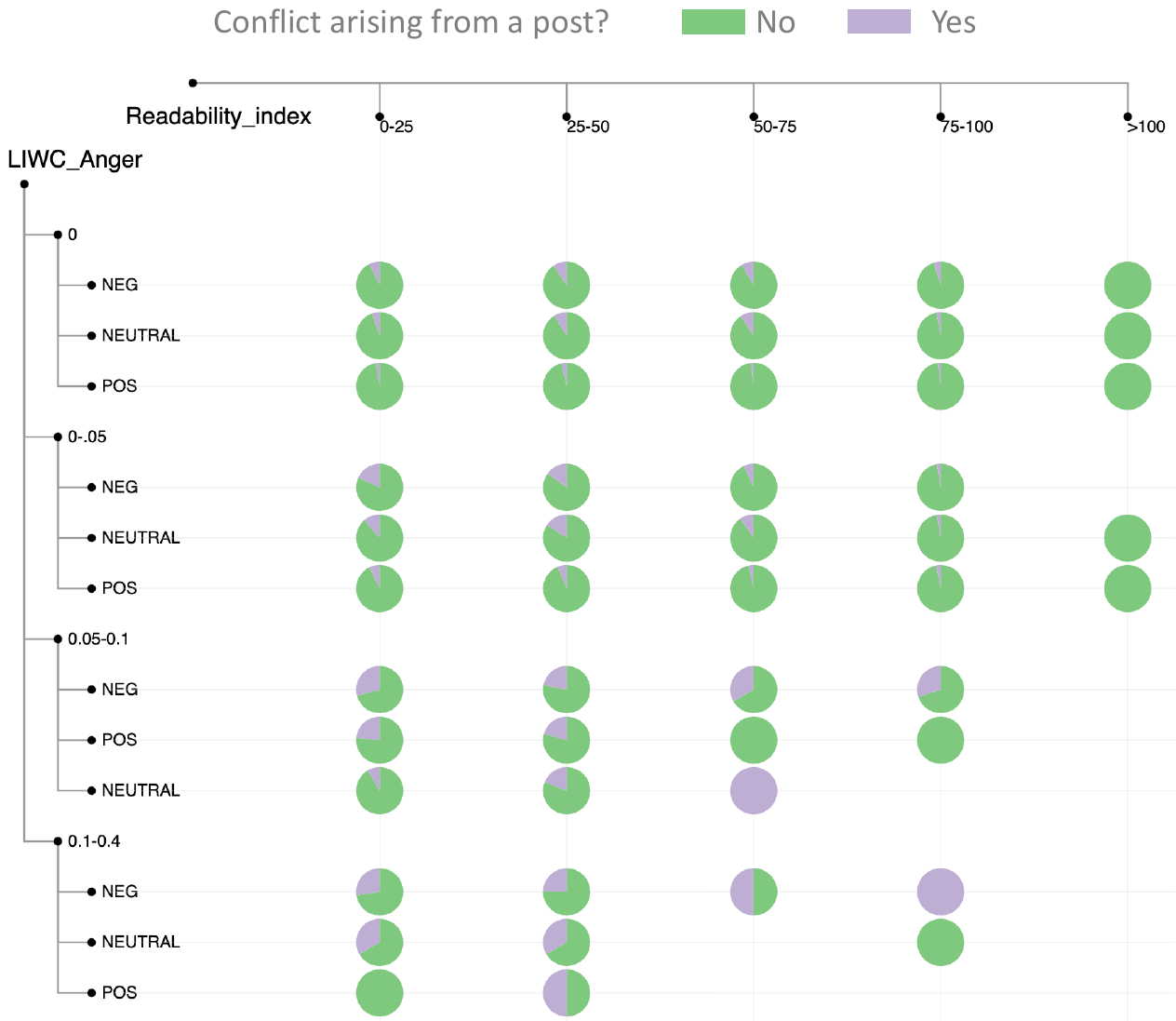}
    \caption{\textbf{Two-dimensional grid view with hierarchical nesting.}
    Analyzing linguistic properties of 300,000 inter-community posts on Reddit~\cite{DBLP:conf/www/KumarHLJ18}.
    Each row in this dataset contains information about a Reddit post from a source to a target community.
    We \faTable~partition the vertical axis by the use of anger-related words, according to Linguistic Inquiry and Word Count (LIWC) and sentiment of the posts, and the horizontal axis by the readability index.
    We \faEye~peek at the variable of interest, whether or not a post  starts a conflict between the source and target community.
    Posts with higher uses of anger-related words give rise to more conflicts among communities.
    Also, posts that are difficult to read (readability index > 100) have less chance of starting a conflict.}
    \label{fig:main_vis}
\end{figure}

\subsection{Visual Representation}

Dataopsy uses a 2D grid view to lay out the supernodes along the horizontal and vertical axis.
Figure~\ref{fig:main_vis} shows an example representation of 300,000 posts among different communities on Reddit~\cite{DBLP:conf/www/KumarHLJ18}.
We describe the details about the representation below.


\subsubsection{Axis Labels}

We used hierarchical grouping to place the labels on the geometric axes.
The order of the variables on the hierarchy depends on the order they were added for \faTable~partitioning by the user.
For example, in Figure~\ref{fig:main_vis}, we first add the readability index and then sentiment on the vertical axis.
We also place variable names on the axis whenever space permits.
On the vertical axis, we only show the first variable name as nested variable names may look indecent. 

\subsubsection{Supernodes}

The visual marks in Dataopsy are typically aggregations of multiple data points, in contrast to the typical one mark per one data.
We call these visual marks \textit{supernodes}, although they may or may not have links depending on whether the domain is a network or not.
In the current implementation of Dataopsy, we represent the supernodes with circles; however, they can be any 2D geometric shapes such as rectangles.

Dataopsy automatically determines the radius of the circles based on Equation~\ref{eq:nodesize}.

\begin{equation}
\begin{split}
    S &= \texttt{min}(width/N_x, height/N_y ) \\
    r &= 
        \begin{cases}
            S, & \text{if } S > \alpha\\
            \alpha, & \text{otherwise}
        \end{cases}
\end{split}
\label{eq:nodesize}
\end{equation}

Here $N_x$ and $N_y$ are the numbers of categories on the horizontal and vertical axes.
$(width, height)$ is the dimension of the SVG, inherited from the card body.
We set $\alpha$, the minimum possible radius, to 5.
When $r = \alpha$, we update the size of the SVG by using Equation~\ref{eq:dimension}.
However, we do not change the size of the card body; instead, we wrap the extended SVG within the card body and provide scrollbars to see the extended contents (see supplement for an example).
This allows us to scale the representation for a large number of categories and avoid visual clutter.

\begin{equation}
\begin{split}
    width &= N_x \cdot r\\
    height &= N_y \cdot r
\end{split}
\label{eq:dimension}
\end{equation}
 
By default, each circle encodes the number of data points in the supernode using a linear color scale.
However, the user can transform the circles into pie charts for seeing distribution along a new dimension using the \faEye~peek operation.
Figure~\ref{fig:main_vis} shows one example of \faEye~peeking, where we see the number of conflicts arising from the Reddit posts in pie charts.
We also place the number of data points at the center of the circles whenever space permits.
For example, Figure~\ref{fig:tablet} and \ref{fig:case_study1} show the numbers in the circles whereas Figure~\ref{fig:main_vis} does not show the numbers due to the small size of the circles.
The user can always see the numbers in a popup when hovering over the circles or pie charts.

\subsubsection{Superlinks}

Similar to the concept of supernodes, we call aggregated links connecting the supernodes \textit{superlinks}. We encode edge weights with the thickness of the links. We used D3's \texttt{arc} function to draw the links.
To reduce visual clutter, we bundle the links and set their default color to light gray with opacity set to 0.3. Despite these design decisions, too many links can create clutter. To avoid that, Superlinks are hidden by default. A user can choose to see all links by toggling the ``Show Links'' option (Figure~\ref{fig:header}b).
Similarly, a user can choose to see direction arrows for the links by toggling ``Show Link Arrows.'' Finally, on hovering over a node, we also highlight links originating and ending at the node with light purple and green colors, respectively (see Figure~\ref{fig:story} and \ref{fig:vispub}).

\subsection{Interactions}
\label{sec:viz_interaction}
We designed fluid interactions to help users perform AQS operations.
There are multiple ways to interact with the visualization in Dataopsy.

The first is to interact with the axis labels.
On hovering over an axis label, Dataopsy highlights all supernodes belonging to that row or column.
Clicking labels will toggle selecting the whole row or column. After selection, a user can use the icon buttons in the card header (Figure~\ref{fig:header}) to \faScissors~prune, \faLightbulbO~project, or \faReorder~pile the selected values. 

Similar to the axis labels, a user can interact with the supernodes directly. On hovering over a node, we highlight the axis labels relevant to the node and  show the number of data points belonging to the node in a popup. If the data contains links, we also show the links originating and ending at the node. On clicking a node, Dataopsy will toggle selecting the data points belonging to the node and allow a user to \faScissors~prune, \faLightbulbO~project, or \faReorder~pile.



\subsection{Implementation Notes}

Dataopsy is currently a web-based prototype.
We used Python running in a \texttt{flask} server as our backend.
We used D3 for rendering the visualization in the frontend.
All user interactions are supported by JavaScript.
The source code and a demo of Dataopsy is available here: \url{https://github.com/tonmoycsedu/Dataopsy}

%% file: content/05-examples.tex
\section{Case Studies and Application Example}

Section 4 demonstrated how AQS can be used to analyze large-scale social media data (Figure~\ref{fig:main_vis}).
In this section, we demonstrate AQS using Dataopsy in four more scenarios: two case studies involving participants and two application examples.

\begin{figure*}
    \centering
    \includegraphics[width=\textwidth]{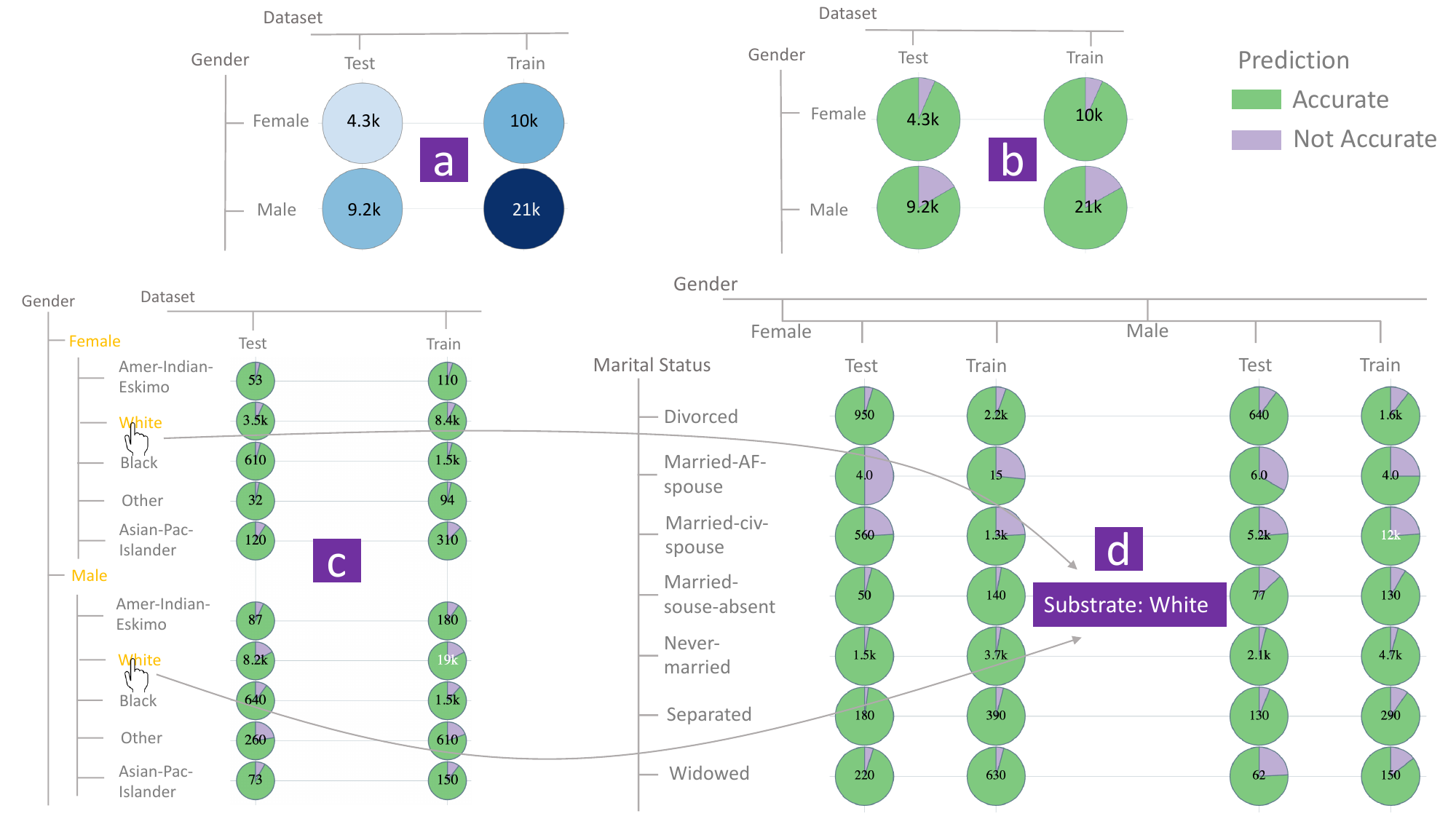}
    \caption{\textbf{Evaluating Fairness in Income Prediction Dataset.}
    (a) Participant (P1) starts by \faTable~partitioning 45,000 data rows into \textsc{Male-Female} and \textsc{Test-Training} subsets.
    (b) Using the \faEye~peek action, P1 evaluates the accuracy of the model on the subsets.
    (c) P1 further \faTable~partitions male and female subsets by race.
    P1 \faLightbulbO~projects \textsc{Female-White} and \textsc{Male-White} (i.e., all \textsc{White} individuals) into a new substrate.
    (d) P1 now \faTable~partitions the new substrate (all \textsc{White}) horizontally by gender and dataset and vertically by marital status.}
    \label{fig:case_study1}
\end{figure*}

\subsection{Case Study: Data Exploration and Fairness Evaluation}

ML and fairness researchers and practitioners often need to ensure how their models perform with respect to sensitive attributes (e.g., gender and race) in the datasets.
This is important since ML models can inherit biases from datasets and propagate the biases in sensitive domains (e.g., loan approval, hiring, and healthcare allocation)~\cite{DBLP:conf/ieeevast/CabreraEHKMC19, DBLP:journals/tvcg/GhaiM23}.

 We worked with two data scientists and fairness researchers to explore how AQS and Dataopsy can be used to evaluate fairness of an ML model.
The first participant (P1) is a male research scientist at a large technology company with a Ph.D.\ in Computer Science and more than 7 years of research experience in data science, visual analytics, and algorithmic fairness.~The second participant (P2) is a female Ph.D.\ student of Computer Science with more than 4 years of research experience in data science and algorithmic fairness.

After a discussion with the participants, we decided to use the Adult Income dataset in this study.
We chose this dataset as it is widely used in the algorithmic fairness and visual analytics literature~\cite{DBLP:conf/cikm/GhaiMM22, DBLP:journals/tvcg/GhaiM23, DBLP:conf/ieeevast/CabreraEHKMC19}.
The dataset contains 45,222 data points where each data point represents a person described by 14 attributes recorded from the U.S.\ 1994 census.
Here, the prediction task is to classify if a person’s income will be greater or less than \$50,000 based on attributes such as age, gender, education, marital status, etc. 

The first author of this paper met with the participants separately over Zoom.~Before the meetings, we asked participants to train a classification model using the dataset.~Both participants used Logistic Regression to train the model.
Their models achieved 86\% accuracy across the training and test sets (70\%-30\% split).
Participants saved the predicted labels and original attributes in a CSV file. 

Each study session started with a training phase where participants explored different features of Dataopsy using a training dataset.
We encouraged participants to ask questions at this stage.
After training, participants uploaded their saved CSV files to Dataopsy and analyzed the model performance.
Participants followed a think-aloud protocol during the study.
The sessions ended with semi-structured interviews focusing on the utility, limitations, and future directions of Dataopsy.

\subsubsection{Results and Feedback}

\paragraph{Evaluating Intersectional Fairness.}

To evaluate the fairness of the trained model, P1 started by \faTable~partitioning the horizontal axis to \textit{train} and \textit{test} sets and the vertical axis to \textit{male} and \textit{female} individuals (Figure~\ref{fig:case_study1}a).
P1 immediately noticed that the dataset is highly skewed towards men. 
Suspecting the skewed dataset might impact accuracy across the subsets, P1 visualized the ratio of the accurate predictions using the \faEye~peek action (Figure~\ref{fig:case_study1}b).
However, the model performed better for females in terms of accuracy.
To investigate further, P1 added race on the vertical axis to \faTable~partition male and female individuals (Figure~\ref{fig:case_study1}c).
P1 noticed that accuracies are consistent across training and test sets for all subsets.
Among the subsets, \textit{Female White} and \textit{Male White} have the highest number of data points.
P1 selected these two subsets and use the \faLightbulbO~projection action to create a new substrate.

P1 then restarted \faTable~partitioning the new substrate, this time by \textit{marital status} on the vertical axis and \textit{gender} and \textit{dataset} on the horizontal axis.
P1 immediately noticed that two subsets, \textit{White Married-AF-spouse} and \textit{White Married-civ-spouse}, were suffering from low accuracies (rows 2 and 3 in Figure~\ref{fig:case_study1}d).
Further, data points pertaining to \textit{White Widowed Male} in the test set had a much lower accuracy than the ones in the training set.
P1 continued this faceted browsing to find under-performing subsets in the dataset.

P2 followed a similar method to evaluate fairness across subsets.
P2 mentioned that Dataopsy allows exploration of the full intersectional space, whereas comparable ML tools often allow only one or two dimensions (e.g., Google Facets).

\paragraph{EDA using Dataopsy.}
Both P1 and P2 thought Dataopsy is a useful tool for exploratory data analysis (EDA).
P2's work requires extensive data analysis, where finding interesting insights often requires hours of coding in computational notebooks.
P2 thought Dataopsy's faceted browsing provides a faster and more structured way to explore a dataset.
P1's work often requires obtaining a balanced dataset across intersectional groups to reduce the chances of biases in model training.
P1 thought Dataopsy is a handy tool to find empty or imbalanced groups or subsets in a dataset.

\paragraph{Recommendation for Dataopsy as a Fairness Tool.}

P1 and P2 provided several recommendations for adapting AQS in a dedicated fairness tool.
One common suggestion was to add functionalities to answer ``why'' a subset of interest suffers from low accuracy after finding the subset.
Such functionalities may include examining individual data rows (P2), observing the distribution of all features inside a subset (P1), and counterfactual analysis (P2).
Another suggestion was to include more metrics to evaluate fairness (e.g., F1 score, and Theil Index).

\begin{figure}
    \centering
    \includegraphics[width=0.99\columnwidth]{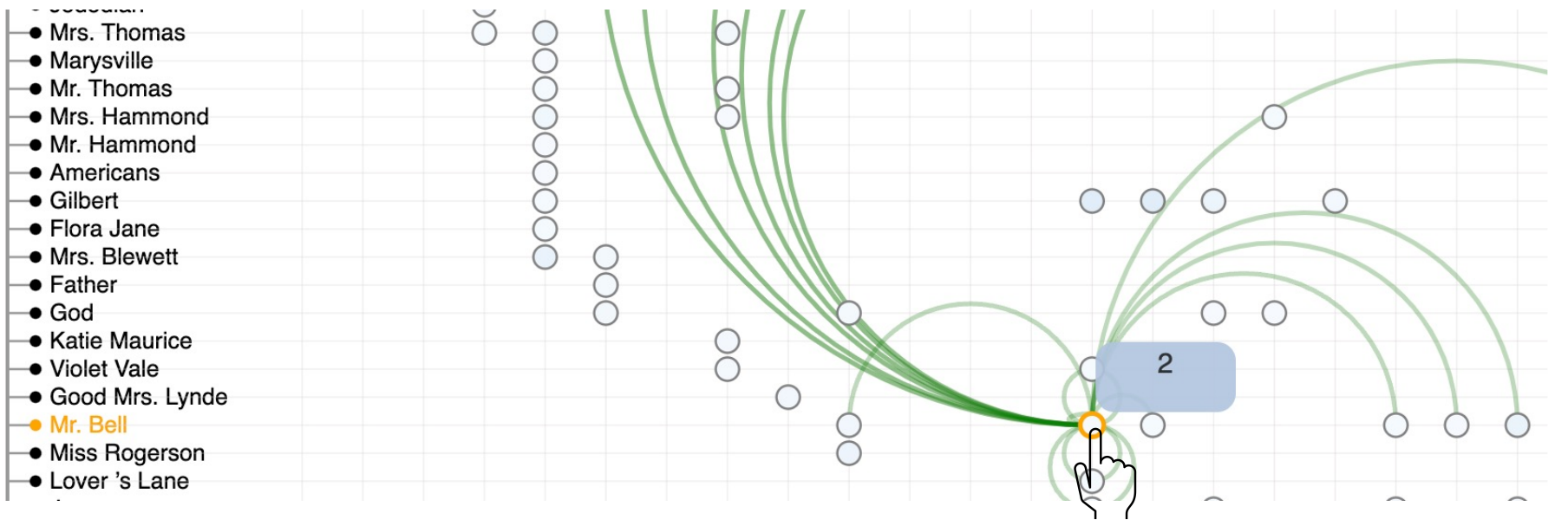}
    \caption{\textbf{Visualizing a story in Dataopsy for writing an adapted screenplay.}
    Here we are seeing a partial view of the story \textit{Anne of Green Gables} by L.\ M.\ Montgomery.
    The horizontal axis represents chapter and the vertical axis represents entities grouped by their type (person, place, etc).
    The color of circles represent the number of mentions, extracted using BookNLP.
    On hover, Dataopsy highlights how an entity (Mr.\ Bell) is connected to other entities in the story.}
    \label{fig:story}
\end{figure}

\begin{figure*}[h]
    \centering
    \includegraphics[width=0.9\textwidth]{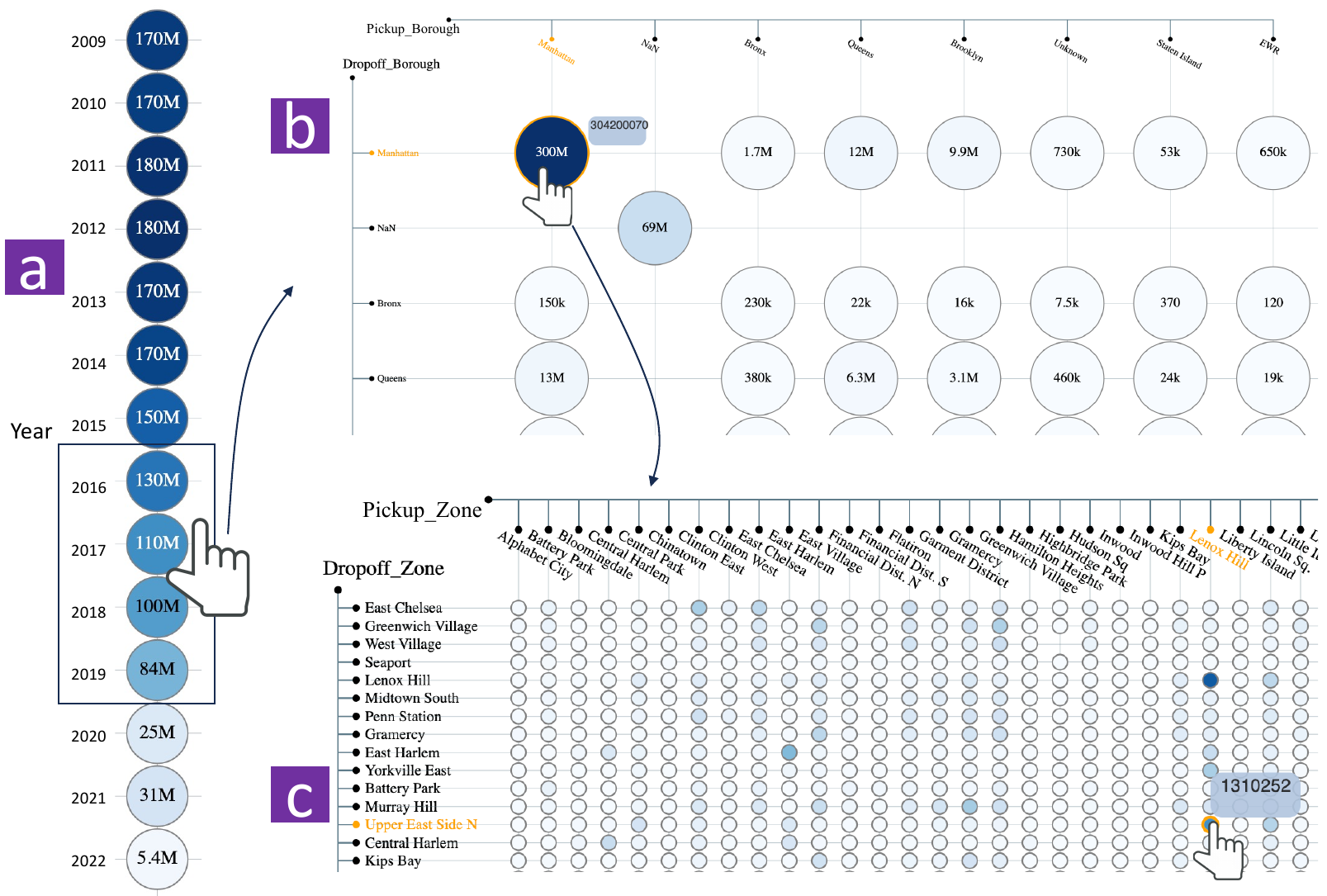}
    \caption{\textbf{Exploring taxi rides in New York City (NYC).}
    (a) We \faTable~partition the whole dataset (around 1.7 billion trips) by \textsc{Year}.
    We select the trips (around 430M) from pre-pandemic years (2016-2019) and \faLightbulbO~project them in a new view. 
    (b) We \faTable~partition the pre-pandemic trips by the pickup and dropoff boroughs.
    Most trips (300M) originated and ended in Manhattan.
    (c) We can further drill down on the Manhattan trips by \faLightbulbO~projecting them on a new view and \faTable~partitioning them by pickup and dropoff zones within Manhattan.
    On hover, we see the number of trips between a hotspot, trips between \textsc{Lenox Hill} and \textsc{Upper East Side North}.}
    \label{fig:nyc}
\end{figure*}

\subsection{Case Study: Planning for an Adapted Screenplay}

In this case study, we demonstrate AQS for a novel application domain: planning for an Adapted Screenplay.
Screenplays dictate the making of films, TV shows, and stage performances.
Screenplays are often adapted, where writers design their stories based on existing texts.
An adapted screenplay is around 120 pages in length whereas a typical novel is around 400-800 pages.
The critical challenge of writing an adapted screenplay is to decide what plotlines, characters, or places from the original story to include in the adapted story.
With consultation from a creative writer (W1), we explore how AQS can be helpful in this scenario.
The writer is a 27 years old female with published articles in their portfolio and a BA degree in English.

For this study, we met with the writer online via Zoom.
Before the meeting, we asked the writer if they have a novel of choice.
The writer chose \textit{Anne of Green Gables} by L.\ M.\ Montgomery, a famous children's book with around 400 pages.
We used BookNLP\footnote{\url{https://github.com/booknlp/booknlp}} to extract different types of entities (Persons, Facilities, and Locations) from the story.
Similar to our previous case studies, the study included a training phase at the start of the session.
We then asked the writer to visualize the collection of entities using Dataopsy and devise a skeleton of the proposed adapted screenplay with the help of the AQS operations.

\subsubsection{Results and Feedback}

\paragraph{Obtaining an Overview from Different Perspectives.}

W1 started exploring the story by \faTable~partitioning the horizontal axis into the 35 chapters of the book.
They then \faTable~partitioned the vertical axis by the type of entities (persons, places, etc.), followed by the actual entities (Figure~\ref{fig:story}).
W1 then quickly hovered over several entities to see how they are connected to each other.

W1 stated that \faTable~partitioning helped them to see the story from different perspectives.
During the session, we noticed W1 continuously changing partitioning order. 
For example, sometimes W1 used entities and chapters to partition the vertical axis linearly.
Other times, W1 used chapters on the horizontal axis and entities on the vertical.

\paragraph{Iterative Pruning and Piling.}

W1 found \faScissors~pruning and \faReorder~piling to be the most useful operations for developing a skeleton for the adapted screenplay.
W1 started by pruning entities with low frequency and dependency in the story. 
Pruning allowed W1 to reduce several plotlines, characters, and places.
W1 also used piling to reduce the size of the story.
For example, after pruning several entities of a chapter, W1 piled (i.e., merged) the chapter with the previous chapter.


\paragraph{Recommendation for Dataopsy as a Writing Support Tool.}

W1 provided several suggestions for adopting AQS and Dataopsy in a writing support tool.
One expected suggestion was to include a text editor and link the text with entities in the visualization.
This would allow writers to see the context in the text and take informed decisions before pruning or piling.
Another suggestion was to include social relations (e.g., brother, mother) as a feature so that writers can decide which characters to prune from a social circle.

\subsection{Example: Understanding Taxi Trips in New York City}

We present an application example on the New York City (NYC) taxi ride dataset~\cite{NYCTaxi} to demonstrate how AQS and Dataopsy scale to large datasets.
The dataset contains every reported trip from 2009 to 2022 in NYC (approximately 1.7 billion trips). 
We chose this application because it is a large dataset (69 GB) with many facets for exploration.

The size of the dataset yields a range of analyses to perform.
For this example, we show how past taxi rides can be analyzed to identify hotspots and devise a policy for allocating taxi cabs in 2022.
Lockdowns and a lack of passengers during the COVID-19 pandemic (2020-2021) heavily disrupted NYC taxi service.\footnote{\url{https://www.cnn.com/2021/01/09/us/yellow-taxi-drivers-new-york-covid/}}
Many taxi drivers changed their profession during this time.
As the world reopened after the pandemic, analyzing past taxi rides can inform the allocation policy of taxis throughout the city.

We used \textsc{Dask}, a Python library for parallel computing, to conduct the backend analysis.
After loading the dataset, we first \faTable~partition the vertical axis by \textsc{Year} (Figure~\ref{fig:nyc}a).
Applying the partition took 30 seconds for Dataopsy.
The number of trips has gradually declined over the years.
As expected, we see a significant drop in 2020-2021.
For demonstration purposes, we only select data from the first two months of 2022.
As taxi rides should presumably return to pre-pandemic status, we select (i.e., \faLightbulbO~Project) the trips from the most recent pre-pandemic years (2016-2019) for further investigation.
We then \faTable~partition the selected trips (430M) by their pickup and dropoff boroughs (Figure~\ref{fig:nyc}b).
As expected, most taxi rides originated and ended in Manhattan.
We can \faLightbulbO~Project the Manhattan trips (300M) and further \faTable~partition these trips by their pickup and dropoff zones within Manhattan (Figure~\ref{fig:nyc}c).
We can now clearly see the hotspots within Manhattan. Note that the default $width$ and $height$ of the SVG are extended using Equation~\ref{eq:dimension} due to the large number of zones. Please see the full screenshot of Figure~\ref{fig:nyc}c in the supplement.
We can further explore the hotspots (e.g., \faLightbulbO~projecting and \faTable~partitioning a hotspot by month).

\begin{figure}[t]
    \centering
    \includegraphics[width=0.8\columnwidth]{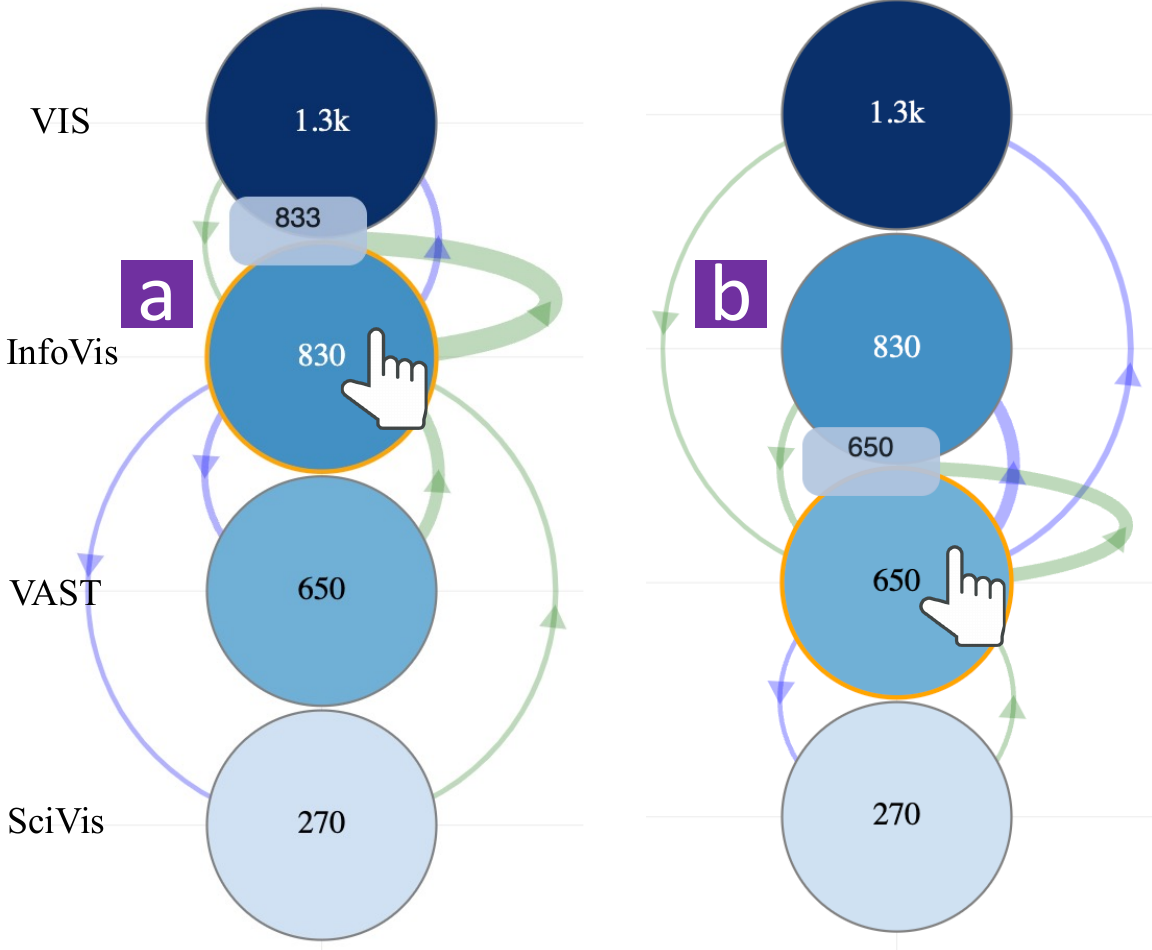}
    \caption{\textbf{Internal citations among the four tracks of IEEE VIS.}
    (a) On hover, light purple links highlight InfoVis papers citing previous papers and light green links show papers citing InfoVis papers.
    (b) Similar analysis for VAST.}
    \label{fig:vispub}
\end{figure}

\subsection{Example: Scientometric Analysis of IEEE VIS Pubs}

Our final example is a scientometric analysis of IEEE VIS publications using the VisPub~\cite{DBLP:journals/tvcg/IsenbergHKIXSSC17, sarvghad2022scientometric} dataset.
This example was designed to show how AQS can be used to analyze multivariate networks.
We can explore this dataset at different levels of aggregation.
For example, Figure~\ref{fig:vispub}a shows a citation network between four conference tracks: Vis, InfoVis, VAST, and SciVis.
On hover, Dataopsy highlights references originated (light purple) at InfoVis and papers citing InfoVis papers (light green).
We noticed that InfoVis papers cite their own papers the most.
VAST papers also cite many papers from VAST, although they cite many InfoVis papers too (Figure~\ref{fig:vispub}b).

%% file: content/06-discussion.tex
\section{Discussion}

Here we discuss design implications, limitations, and future work relevant to AQS and Dataopsy.

\paragraph{Changing Perspective with Partitioning.}

We noticed that participants often used different combinations to partition data, even within the same session.
One key observation is that the order of partitioning can change the perspective even if the underlying data is the same. 
It can potentially affect what information or insights people see first.
This phenomenon gives rise to several interesting questions for the VIS community such as \textit{How exactly does the order of pivoting and partitioning impact the data exploration process?} and \textit{Is there an optimum order to hierarchically nest the partitions?}
Prior work on finding optimum ordering for parallel coordinates are inspiring in this case~\cite{DBLP:conf/apvis/ZhangMM12, DBLP:journals/ivs/RosarioRBWH04}.

\paragraph{Designing and Evaluating Fluid Interaction.}

Our case studies show promises for interaction design for exploring multivariate data.
The supported interactions in AQS ($\mathbb{P}^6$) are larger than in a typical visualization system.
Although AQS was not formally evaluated in the case studies, participants used praises such as ``cool,'' ``nice,'' and ``wow'' to describe the usability of Dataopsy.
Our future work will focus on evaluating Dataopsy in comparison to similar methods (e.g., Google Facets~\cite{Facets2017, Wexler2017}).
We can ask users to find answers to queries (e.g., What percentage of white, married, and female individuals were accurately labeled by the model?) and measure efficiency by counting and comparing the number of steps taken to answer the queries using different methods.
We can further use NASA-TLX~\cite{hart2006nasa} and SUS~\cite{brooke1996} to evaluate the perceived workload and usability of Dataopsy.

\paragraph{Trade-offs between Aggregated and Unit Representation.}

AQS is a top-down technique where the exploration starts with a single mark aggregating all data items. 
This strong aggregation enables us to scale analysis to large datasets.
However, compared to unit visualizations, there are fewer chances of serendipitous findings using our approach.
Due to the lack of an overview, users need to have prior knowledge and hypotheses to construct the queries.
We can partially address this limitation by introducing a recommender system that can recommend subsets using anomaly detection algorithms and prior user interactions~\cite{DBLP:conf/ieeevast/CabreraEHKMC19, DBLP:journals/jourTVCG/pandey2023}.

\paragraph{Scalability.}

As a theoretical concept, AQS is scalable to any number of data points.
However, as an early prototype, Dataopsy currently lacks a few engineering features for handling ``really big'' datasets.
For example, despite using parallel computing, for the NYC taxi ride dataset, on average, it took 30 seconds for Dataopsy to respond to the data operations (e.g., partitioning).
There are established methods to handle such large datasets in visualization displays~\cite{DBLP:journals/cgf/LiuJH13, DBLP:journals/tvcg/PahinsSSC17, DBLP:conf/chi/MoritzHH19}, which could further improve response time.

\paragraph{Visualizing Quantitative Values.}

Dataopsy has limitations for analyzing quantitative values.
For example, when visualizing a quantitative attribute using \faEye~peeking, Dataopsy uses binning and categorical color scales to show the distribution in a pie chart, which can be challenging to decode. 
One option is to extend the supported visual mark type (e.g., histogram in rectangles) to resolve this issue.
We will follow the example of prior work such as Polaris~\cite{DBLP:conf/infovis/StolteH00} to integrate this feature into our tool.
Another relevant problem with \faEye~peeking is that it becomes difficult to measure the size of the nodes (i.e., cardinality) without the color saturation (Figure~\ref{fig:main_vis}).
A solution could be using circular curves along the circles/pies to indicate the size.
Another possible solution is using varying circle sizes to represent cardinality.

\paragraph{Adopting Aggregate Query Sculpting.}

We recommend that practitioners and researchers adopt AQS if the following conditions are met:

\begin{itemize}
\item The data has a sufficient amount of facets (>=2).
\item The number of data points is too many for unit visualization. 
\item The goal is to obtain higher-level insights and patterns rather than finding lower-level similarities between individual data points.
(For example, AQS may not be feasible for finding clusters in an embedding space.)
\item Domain-specific functionalities for the application are easy to integrate with AQS.
\end{itemize}

%% file: content/07-conclusion.tex
\section{Conclusion}
We have presented Aggregate Query Sculpting (AQS), a novel interaction technique for visualizing and exploring multivariate data. The goal of our work was to solve challenges for large-scale data containing many attributes. Visualizing such datasets using unit visualizations (e.g., scatter plots) often results in visual clutter and inelegant representation. We propose aggregation to be key for solving this issue.
As a born scalable technique, AQS initially aggregates all data points into a single visual mark, a supernode. From there, AQS provides six operations, abbreviated as $\mathbb{P}^6$, to iteratively sculpt the data to a desired form. Based on the concept of AQS, we developed Dataopsy, a prototype tool for exploring multivariate data. Dataopsy is equipped to analyze multivariate data from versatile domains. We hope our work will motivate future research for designing visualization that is equipped to manage large-scale data, yet easy to explore.
